\documentclass[]{article}
\makeatletter\if@twocolumn\PassOptionsToPackage{switch}{lineno}\else\fi\makeatother

\usepackage{amsfonts,amssymb,amsbsy,latexsym,amsmath,tabulary,graphicx,times,caption,fancyhdr}
\usepackage[utf8]{inputenc}
\usepackage{url,multirow,morefloats,floatflt,cancel,tfrupee}
\makeatletter

\AtBeginDocument{\@ifpackageloaded{textcomp}{}{\usepackage{textcomp}}}
\makeatother
\usepackage{colortbl}
\usepackage{xcolor}
\usepackage{pifont}
\usepackage[nointegrals]{wasysym}
\makeatletter

\def\mcWidth#1{\csname TY@F#1\endcsname+\tabcolsep}

\def\cAlignHack{\rightskip\@flushglue\leftskip\@flushglue\parindent\z@\parfillskip\z@skip}
\def\rAlignHack{\rightskip\z@skip\leftskip\@flushglue \parindent\z@\parfillskip\z@skip}

\@ifundefined{etal}{}{}

\usepackage{ifxetex}
\ifxetex\else\if@twocolumn\@ifpackageloaded{stfloats}{}{\usepackage{dblfloatfix}}\fi\fi

\AtBeginDocument{
\expandafter\ifx\csname eqalign\endcsname\relax
\def\eqalign#1{\null\vcenter{\def\\{\cr}\openup\jot\m@th
  \ialign{\strut$\displaystyle{##}$\hfil&$\displaystyle{{}##}$\hfil
      \crcr#1\crcr}}\,}
\fi
}

\AtBeginDocument{%
  \@ifpackageloaded{endfloat}%
   {\renewcommand\efloat@iwrite[1]{\immediate\expandafter\protected@write\csname efloat@post#1\endcsname{}}}{\newif\ifefloat@tables}%
}%

\def\BreakURLText#1{\@tfor\brk@tempa:=#1\do{\brk@tempa\hskip0pt}}
\let\lt=<
\let\gt=>
\def\processVert{\ifmmode|\else\textbar\fi}

\@ifundefined{subparagraph}{
\def\subparagraph{\@startsection{paragraph}{5}{2\parindent}{0ex plus 0.1ex minus 0.1ex}%
{0ex}{\normalfont\small\itshape}}%
}{}

\newcommand\role[1]{\unskip}
\newcommand\aucollab[1]{\unskip}
  
\@ifundefined{tsGraphicsScaleX}{\gdef\tsGraphicsScaleX{1}}{}
\@ifundefined{tsGraphicsScaleY}{\gdef\tsGraphicsScaleY{.9}}{}
\def\checkGraphicsWidth{\ifdim\Gin@nat@width>\linewidth
	\tsGraphicsScaleX\linewidth\else\Gin@nat@width\fi}

\def\checkGraphicsHeight{\ifdim\Gin@nat@height>.9\textheight
	\tsGraphicsScaleY\textheight\else\Gin@nat@height\fi}

\def\fixFloatSize#1{}%\@ifundefined{processdelayedfloats}{\setbox0=\hbox{\includegraphics{#1}}\ifnum\wd0<\columnwidth\relax\renewenvironment{figure*}{\begin{figure}}{\end{figure}}\fi}{}}
\let\ts@includegraphics\includegraphics

\def\inlinegraphic[#1]#2{{\edef\@tempa{#1}\edef\baseline@shift{\ifx\@tempa\@empty0\else#1\fi}\edef\tempZ{\the\numexpr(\numexpr(\baseline@shift*\f@size/100))}\protect\raisebox{\tempZ pt}{\ts@includegraphics{#2}}}}

\AtBeginDocument{\def\includegraphics{\@ifnextchar[{\ts@includegraphics}{\ts@includegraphics[width=\checkGraphicsWidth,height=\checkGraphicsHeight,keepaspectratio]}}}

\DeclareMathAlphabet{\mathpzc}{OT1}{pzc}{m}{it}

\def\URL#1#2{\@ifundefined{href}{#2}{\href{#1}{#2}}}

\def\UrlOrds{\do\*\do\-\do\~\do\'\do\"\do\-}%
\g@addto@macro{\UrlBreaks}{\UrlOrds}

\edef\fntEncoding{\f@encoding}

\makeatother

\newif\ifmultipleabstract\multipleabstractfalse%
\makeatletter

\def\wileyIndent{1pt}
\usepackage[paperheight=10in,paperwidth=6.5in,margin=2cm,headsep=.5cm,top=2.5cm,headheight=1cm]{geometry}

\renewenvironment{abstract}
{\vspace*{-1pc}\trivlist\item[]\leftskip\wileyIndent\hrulefill\par\vskip4pt\noindent\textbf{\abstractname}\mbox{\null}\\}{\par\noindent\hrulefill\endtrivlist}

\usepackage[]{footmisc}

\def\author#1{\gdef\@author{\hskip-\dimexpr(\tabcolsep)\hskip\wileyIndent\parbox{\dimexpr\textwidth-\wileyIndent}{\centering\bfseries#1}}}

\def\title#1{\linespread{1}\gdef\@title{\centering\bfseries\ifx\@articleType\@empty\else\@articleType\\\fi#1}}

\let\@articleType\@empty \def\articletype#1{\gdef\@articleType{{\normalfont\itshape#1}}}

\AtBeginDocument{\fancypagestyle{headings}{\fancyhf{}\fancyhead[C]{\RunningHead}\fancyhead[R]{\thepage}}\pagestyle{headings}}

 \def\audegree#1{}

\date{}

\makeatother

\usepackage[T1]{fontenc}
\makeatother
\usepackage[numbers,sort&compress]{natbib}

\def\thanksspace{{\phantom{\textsuperscript{\thefootnote}}}}

\usepackage{amsmath,amssymb,amsfonts}
\usepackage{algorithmic}
\usepackage{graphicx}
\usepackage{textcomp}

\usepackage{makecell}
\usepackage{multirow}

\usepackage{geometry}
\begin{document}

\title{A convolutional neural network for teeth margin detection on 3-dimensional dental meshes}
\author{Hu~Chen\textsuperscript{1,4}\thanks{Corresponding author.}\thanksspace \thanks{E-mail:                     
                    ccen@bjmu.edu.cn}{\thanksspace}\thanks{Hu Chen and Hong Li contributed equally}\thanksspace\space Hong~Li\textsuperscript{2,4}\space Bifu~Hu\textsuperscript{3,4}\space Kenan~Ma\textsuperscript{1,4}\space and Yuchun~Sun\textsuperscript{1,4}~\\[-3pt]\normalsize\normalfont  \itshape ~\\
\textsuperscript{1}{Center of Digital Dentistry, Department of Prosthodontics, National Engineering Laboratory for Digital and material technology of stomatology,  Research Center of Engineering and Technology for Digital Dentistry, Peking University School and Hospital of Stomatology, Bejing, 100081 China}~\\
\textsuperscript{2}{First Clinical Division, Peking University School and Hospital of Stomatology, Bejing, 100034 China}~\\
\textsuperscript{3}{School of Mechanical Engineering and Automation, Beihang University, Bejing, 100191 China}~\\
\textsuperscript{4}{Shanxi Province Key Laboratory of Oral
	Diseases Prevention and New Materials, Shanxi Medical University, Shanxi, 030001, China}}

\def\RunningHead{Mesh CNN for teeth margin detection}\def\RunningAuthor{Hu and Hong}

\maketitle

\begin{abstract}
We proposed a convolutional neural network for vertex classification on 3-dimensional dental meshes, and used it to detect teeth margins. An expanding layer was constructed to collect statistic values of neighbor vertex features and compute new features for each vertex with convolutional neural networks. An end-to-end neural network was proposed to take vertex features, including coordinates, curvatures and distance, as input and output each vertex classification label. Several network structures with different parameters of expanding layers and a base line network without expanding layers were designed and trained by 1156 dental meshes. The accuracy, recall and precision were validated on 145 dental meshes to rate the best network structures, which were finally tested on another 144 dental meshes. All networks with our expanding layers performed better than baseline, and the best one achieved an accuracy of 0.877 both on validation dataset and test dataset. \def\keywordstitle{Keywords}

\end{abstract}
    
\section{Introduction}
With development of digital dentistry, optical scanning of dental models became essential for computer aided design (CAD) and computer aided manufacture (CAM) of dental prosthetics \cite{PradeepNarasimman-4025,ZandinejadLin-4026,RunkelGuth-4027,JodaFerrari-4028} and for digital orthodontics \cite{IgnaTodea-4029}, for example, digital indirect bonding technique \cite{DuarteGribel-4030}, production of vacuum formed invisible aligners \cite{Andronic-4031}. The optical scanners, based on principle of triangle measuring method, acquire model surface point cloud data and convert them to mesh (always triangular mesh). The mesh data contains vertex coordinates and their connections (edges and facets). Teeth in the mesh should be segmented before design of dental prosthetics or make virtual orthodontic setup, which may take enormous and strenuous manual labor.

There are several kinds of classical algorithm (without machine learning) for mesh segmentation\cite{Shamir-4032,RodriguesMorgado-4033,TheologouPratikakis-4034}, such as region growing\cite{RodriguesMorgado-4035,HolzBehnke-4036,BergamascoAlbarelli-4037,LavoueDupont-4038}, watershed segmentation\cite{BenjaminPolk-4039,ChenGeorganas-4040,ZuckerbergerTal-4041,A.R.-4042},contour-based mesh segmentation\cite{RodriguesMorgado-3971}, hierarchical clustering\cite{ZhangLi-4043,YanWang-4044,Reuter-4045,De_GoesGoldenstein-4046,InoueItoh-4047}, iterative clustering\cite{ZhangZheng-4048,Wu_Leif_Kobbelt-4049,JuliusKraevoy-4050,Cohen-SteinerAlliez-4051,ShlafmanTal-4052}, spectral analysis\cite{LiuZhang-4053,KarniGotsman-4054} and implicit methods (define the boundaries between sub-meshes)\cite{WangLu-4055,BenhabilesLavoue-4056,ZhengTai-4057,GolovinskiyFunkhouser-4058,SchreinerAsirvatham-4059}, which are not reliable in complex cases and not easy to extend. With the development of artificial intelligence, machine learning methods\cite{GeorgeXie-4060,QiLi-4061,KalogerakisAverkiou-4062,YiSu-4063,QiSu-4064,P.I.-4065,GuoZou-4066} were applied for mesh segmentation. Some researches collected hundreds of vertex or facet features to do classification on support vector machine or neural network\cite{GeorgeXie-4060,GuoZou-4066}, while the features are complex to define and calculate. Several recent researches aimed to take mesh as raw input and do segmentation work by converting to voxel for 3-dimentional (3D) convolutional neural network (CNN)\cite{WuSong-4067,D.S.-4068,QiSu-4069}, or projection to 2-dimentiional (2D) images for 2D CNN\cite{KalogerakisAverkiou-4062,LeBui-4070}, or direct point clouds deep learning\cite{QiLi-4061,QiSu-4064}. However, the innate vertex connections in the mesh were ignored in those methods, which is very important for the cognition of mesh shape and structure.
Here we propose an artificial neural network named mesh CNN, where only a few features for each vertex need to be precalculated, and the mesh vertex feature matrix serve as raw input without voxelization or projection. In this model, the topological connections between mesh vertexes were utilized to produce an adjacent matrix, with which neighborhood features for each vertex were involved for calculate, expanding receptive filed iteratively, and finally, vertex features after convolution were sent to soft-max layer. We challenged clinical dentition plaster casts segmentations and achieved good performances.

\begin{figure*}[ht]
	\centering
	\includegraphics[scale=0.095]{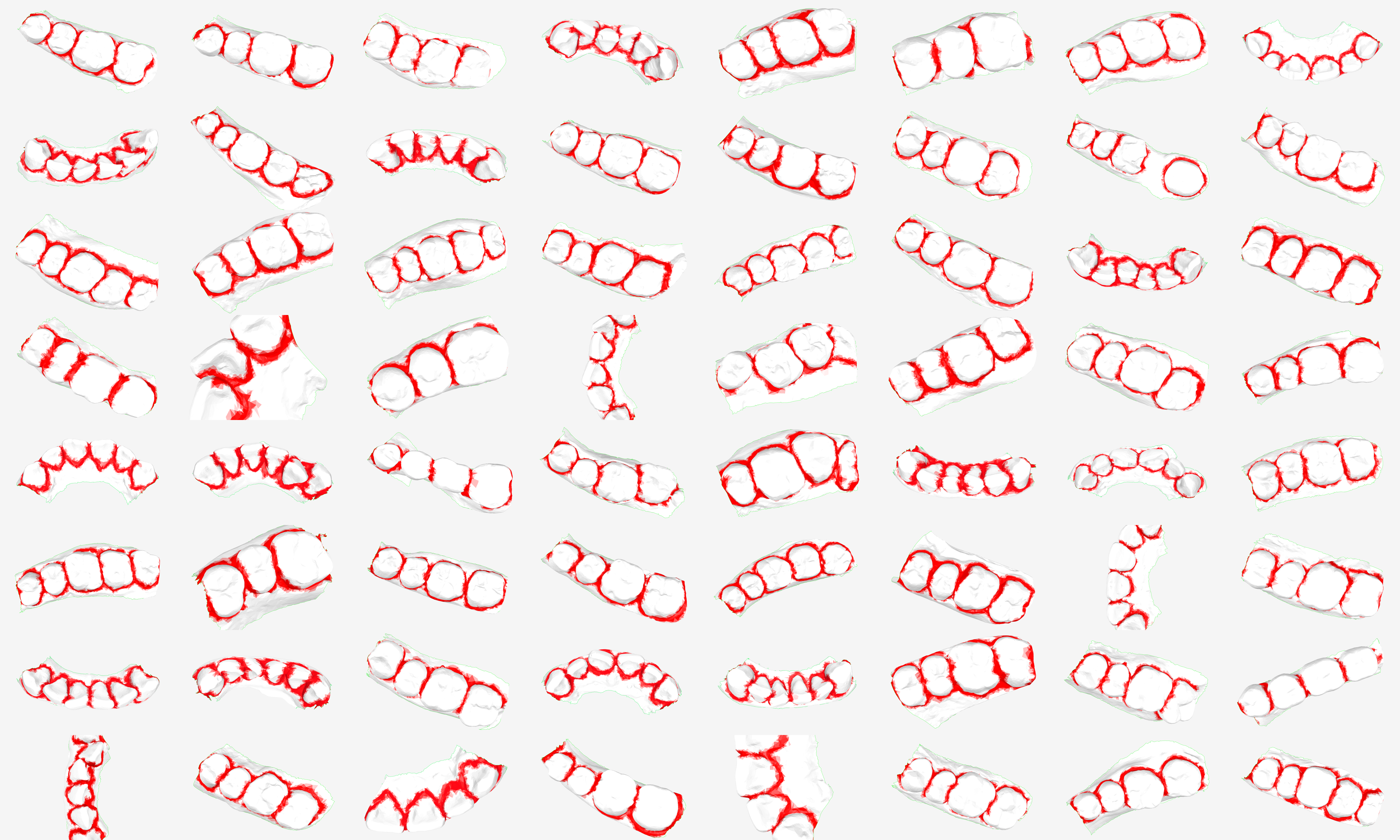}
	\caption{Annotations of teeth margins on dentition meshes, occlusal view.}
	\label{fig:annos}
\end{figure*}

In recent years, machine learning has been widely applicated in the field of mesh segmentations. Kalogerakis et al.\cite{KalogerakisHertzmann-4071} presented a data-driven approach to simultaneous segmentation and labeling of parts in 3D meshes based on Conditional Random Field model, where they achieved improvement on the Princeton Segmentation Benchmark. Guo et al.\cite{GuoZou-4066} trained CNNs by a large pool of classical mesh facet geometric features, to initialize a label vector for each triangle indicating its probabilities of belonging to various object parts, but connections between mesh facets were unable to be concerned in these CNNs.Xu et al.\cite{X.C.-4129} constructed 2-level hierarchical CNNs structure for tooth segmentation, consuming a set of geometry features (600-dimension) as face feature representations to train the networks. George et al.\cite{GeorgeXie-4060} used multi-scale features calculated as the mean of a set of neighboring facets to train multi-branch 1D CNNs, thus to classify a facet by its multi scale local area features. However, hundreds of geometric and contextual features need to be defined and calculated aforehand for their networks.

Wu et al.\cite{WuSong-4067} proposed 3D shapeNets, which represent a geometric 3D shape as a probability distribution of binary variables on a 3D voxel grid based on Convolutional Deep Belief Network. Except transferring to voxel, multi-view images were also used to represent 3D mesh geometric information. Le et al.\cite{LeBui-4070} introduced a multi-view recurrent neural network approach for 3D mesh segmentation, where the convolutional neural networks and long short term memory were combined to generate the edge probability feature map and their correlations across different views. Qi et al.\cite{QiSu-4069} combined the volumetric representations and multi-view representations, with introducing of architectures of volumetric CNNs, as well as introducing of multi-resolution filtering in 3D into multi-view CNNs, to provide better methods available for object classification on 3D mesh data. However, transferring mesh to voxel grids of different view or to images may produce unnecessarily intermediate steps and raise issues. PointNet\cite{QiSu-4064} was designed by Qi et al. for applications of 3D object classification, part segmentation, and scene semantic parsing, which directly consumes point clouds, and respects the permutation invariance of points in the input. PointNet was further developed to be PointNet++\cite{QiLi-4061}, which applies PointNet recursively on a nested partitioning of the input pointset, to learn deep point set features efficiently and robustly. Different from point clouds, the teeth mesh data have vertex connections represented by the edges of facets. If treated as point clouds, the geometric connections between mesh vertex, which is very important to retrieve the mesh topological structures, will lost. Thus, in this research, we proposed an architecture of a neural network named mesh CNN that exploits the edge connection information to compute local area features on mesh data, and finally implement the vertex classification.

\section{Proposed Approach}
We constructed a neural network for vertex detection, where an expanding layer was proposed, which utilized the connection information of the triangle meshes to generate links between vertexes, concatenating features of one vertex to that of its adjacent vertexes. CNN was applied to the concatenated features, after serval repeat of the expanding layer and CNN layer architectures, probability of labels for each vertex was finally output. The details of expanding layer are described below.
In a mesh with $\mathit{n}$ vertexes and $\mathit{N_{e}}$ edges, all vertexes ($\mathit{V}$) and edges ($\mathit{E}$) can be defined as:
\begin{equation}
V= \{ v|v= \left ( x_{i},y_{i},z_{i}  \right) \in  \mathbb{R}^3, i=1,2,\cdot \cdot \cdot ,n \}
\end{equation}
\begin{equation}
E=\{e|e=\{v_{i1},v_{i2}\},i=1,2,\cdot \cdot \cdot ,N_{e}\}
\end{equation}

For any vertex $v_{i}$, the vertexes directly connected to $v_{i}$ are those on the same edge with $v_{i}$, which can be defined as:
\begin{equation}
\boldsymbol{N}\left ( v_{i} \right )=\left \{ v|\left \{ v,v_{i} \right \}\in E \right \}
\end{equation}

Further, for $\forall~ V_{s}\subseteq V$,
\begin{equation}
\boldsymbol{N}\left ( V_{s} \right )=\left \{ v|\left \{ v,v_{i} \right \}\in E,v_{i} \in V_{s} \right \}
\end{equation}

In this research, we define ring 0 neighbor vertexes of $v_{i}$  as $v_{i}$ its self:
\begin{equation}
\boldsymbol{N}_{0}\left ( v_{i} \right )=v_{i}
\end{equation}

Ring-1 neighbor vertexes of $v_{i}$ as vertexes directly connected to $v_{i}$:
\begin{equation}
\boldsymbol{N}_{1}\left ( v_{i} \right )=\boldsymbol{N}\left ( v_{i} \right )
\end{equation}

Ring-2 neighbor vertexes of $v_{i}$ as vertexes directly connected to ring 1 neighbor vertexes of $v_{i}$, excluding ring 1 neighbor vertexes and $v_{i}$:
\begin{equation}
\boldsymbol{N}_{2}\left (v_{i}  \right )=\boldsymbol{N}\left (\boldsymbol{N}_{1}\left (v_{i}  \right )  \right )\setminus \left (\boldsymbol{N}_{1}(v_{i})\cup \boldsymbol{N}_{0}\left ( v_{i} \right )  \right )
\end{equation}

For $n\geq 2$, ring $n$ neighbor vertexes of $v_{i}$ as vertexes directly connected to ring $\left(n-1\right)$ neighbor vertexes of $v_{i}$, excluding ring $\left(n-1\right)$ neighbor vertexes and inner vertexes:
\begin{equation}
\begin{aligned}
&\boldsymbol{N}_{n}\left (v_{i}  \right )=\\ &\boldsymbol{N}\left (\boldsymbol{N}_{n-1}\left (v_{i}  \right )  \right )\setminus \left (\boldsymbol{N}_{n-1}(v_{i})\cup \boldsymbol{N}_{n-2}\left ( v_{i} \right ) \cup \cdot \cdot \cdot \cup \boldsymbol{N}_{0}\left ( v_{i} \right ) \right )
\end{aligned}
\end{equation}

It can be easily inferred that ring n neighbor vertexes will not directly connect to ring $n-2$ and inner vertexes, that is:
\begin{equation}
\boldsymbol{N}\left (\boldsymbol{N}_{n-1}\left (v_{i}  \right )  \right )\cap \left (\boldsymbol{N}_{n-3}(v_{i}) \cup \cdot \cdot \cdot \cup \boldsymbol{N}_{0}\left ( v_{i} \right ) \right )=\phi 
\end{equation}

So, the equation 8 can be simplified as:
\begin{equation}
\boldsymbol{N}_{n}\left (v_{i}  \right )=\boldsymbol{N}\left (\boldsymbol{N}_{n-1}\left (v_{i}  \right )  \right )\setminus \left (\boldsymbol{N}_{n-1}(v_{i})\cup \boldsymbol{N}_{n-2}\left ( v_{i} \right ) \right )
\end{equation}

For a vertex $v_{i}$, there might be several features, such as coordinates $\left(x,y,z \right)$, curvatures, distances, which can be presented as:
\begin{equation}
\mathbf{Ft}\left ( v_{i} \right )=\begin{bmatrix}
ft_{1}\left ( v_{i} \right ) & ft_{2}\left ( v_{i} \right ) & \cdot \cdot \cdot & ft_{m}\left ( v_{i} \right ) 
\end{bmatrix}
\end{equation}

For $\forall~V_{s} \subseteq V$, we define the mean features of vertexes in $V_{s}$ as:
\begin{equation}
\begin{split}
&\mathbf{MFt}\left ( V_{s} \right )=\\
&\begin{bmatrix}
\frac{1}{s}\sum_{v \in V_{s}}ft_{1}\left ( v \right ) & \frac{1}{s}\sum_{v \in V_{s}}ft_{2}\left ( v \right ) & \cdot \cdot \cdot & \frac{1}{s}\sum_{v \in V_{s}}ft_{m}\left ( v \right )
\end{bmatrix}
\end{split}
\end{equation}
where $s$ is the number of vertexes in $V_{s}$.
Consider input of a mesh matrix with all $n$ vertexes, each vertex with $m$ features:
\begin{equation}
\mathbf{X}=\begin{bmatrix}
\mathbf{Ft}\left ( v_{1} \right )\\ 
\mathbf{Ft}\left ( v_{2} \right )\\ 
\cdot \cdot \cdot \\ 
\mathbf{Ft}\left ( v_{n} \right )
\end{bmatrix}
\end{equation}

We can calculate its mean ring $k$ neighbor features:
\begin{equation}
\mathbf{X}^{k}=\begin{bmatrix}
\mathbf{MFt}\left ( \boldsymbol{N}_{k}\left ( v_{1} \right ) \right )\\ 
\mathbf{MFt}\left ( \boldsymbol{N}_{k}\left ( v_{2} \right ) \right )\\ 
\cdot \cdot \cdot \\ 
\mathbf{MFt}\left ( \boldsymbol{N}_{k}\left ( v_{n} \right ) \right )
\end{bmatrix}
\end{equation}
where $\mathbf{X}^{k}$ contain some context information of mesh connections, and it can be easily inferred that $\mathbf{X}^{0}=\mathbf{X}$. The expanding layer, take mesh $n \times m$ feature matrix $\left( \mathbf{X} \right) $ as input, and output an augmented matrix with means of several ring neighbor features:

\begin{equation}
\label{eq_expanding_layer}
\begin{split}
&Expand^{0,1,\cdot \cdot \cdot,k}\left(\mathbf{X} \right )=\begin{bmatrix}
\left ( \mathbf{X}^{0}  \right )^{T} \\ 
\left ( \mathbf{X}^{1}  \right )^{T}\\ 
\cdot \cdot \cdot \\ 
\left ( \mathbf{X}^{k}  \right )^{T}
\end{bmatrix}
=\\
&\begin{bmatrix}
\mathbf{MFt}\left ( \boldsymbol{N}_{0}\left ( v_{1} \right ) \right ) & \mathbf{MFt}\left ( \boldsymbol{N}_{0}\left ( v_{2} \right ) \right ) & \cdot \cdot \cdot & \mathbf{MFt}\left ( \boldsymbol{N}_{0}\left ( v_{n} \right ) \right ) \\ 
\mathbf{MFt}\left ( \boldsymbol{N}_{1}\left ( v_{1} \right ) \right ) & \mathbf{MFt}\left ( \boldsymbol{N}_{1}\left ( v_{2} \right ) \right ) & \cdot \cdot \cdot & \mathbf{MFt}\left ( \boldsymbol{N}_{1}\left ( v_{n} \right ) \right ) \\ 
\cdot \cdot \cdot & \cdot \cdot \cdot & \cdot \cdot \cdot & \cdot \cdot \cdot \\ 
\mathbf{MFt}\left ( \boldsymbol{N}_{k}\left ( v_{1} \right ) \right ) & \mathbf{MFt}\left ( \boldsymbol{N}_{k}\left ( v_{2} \right ) \right ) & \cdot \cdot \cdot & \mathbf{MFt}\left ( \boldsymbol{N}_{k}\left ( v_{n} \right ) \right ) 
\end{bmatrix}
\end{split}
\end{equation}

Here is an instance of matrix computation algorithm to implement the $Expand$ function. Firstly, we constructed an adjacent matrix $\mathbf{A}$, which represent the connection relationship between mesh vertexes:
\begin{equation}
\mathbf{A}_{k,i,n}=  \left\{\begin{matrix}
1,~ v_{n} \in \boldsymbol{N}_{k}\left ( v_{i} \right )\\ 
0,~ v_{n} \notin \boldsymbol{N}_{k}\left ( v_{i} \right )
\end{matrix}\right.
\end{equation}

Secondly, construct a new matrix $\mathbf{B}$ to collect ring neighbor features:
\begin{equation}
\mathbf{B}_{k,i,m,n}=\mathbf{A}_{k,i,n}\ast \mathbf{X}_{n,m}
\end{equation}

Finally, construct another new matrix $\mathbf{C}$ computing means of ring neighbor features:
\begin{equation}
\mathbf{C}_{k,i,m}=\frac{\sum_{n}\mathbf{B}_{k,i,m,n}}{\sum_{n}\mathbf{A}_{k,i,n}}
\end{equation}

Then, the constructed matrix $\mathbf{C}$ will be equal to output of expanding layer, as defined in equation  \ref{eq_expanding_layer}:
\begin{equation}
Expand^{0,1,\cdot \cdot \cdot,k}\left ( \mathbf{X} \right )=\mathbf{C}
\end{equation}

The input features can also be expanded to non-continuous ring number neighbor features, of which the result is a slice of $\mathbf{C}$, for example:
\begin{equation}
Expand^{0,2,4}\left ( \mathbf{X} \right )=\mathbf{C}_{\left [ 0,2,4 \right ],:,:}
\end{equation}
\begin{equation}
Expand^{0,4,8}\left ( \mathbf{X} \right )=\mathbf{C}_{\left [ 0,4,8 \right ],:,:}
\end{equation}

\begin{figure*}[htbp]
	\centering
	\includegraphics[scale=0.17]{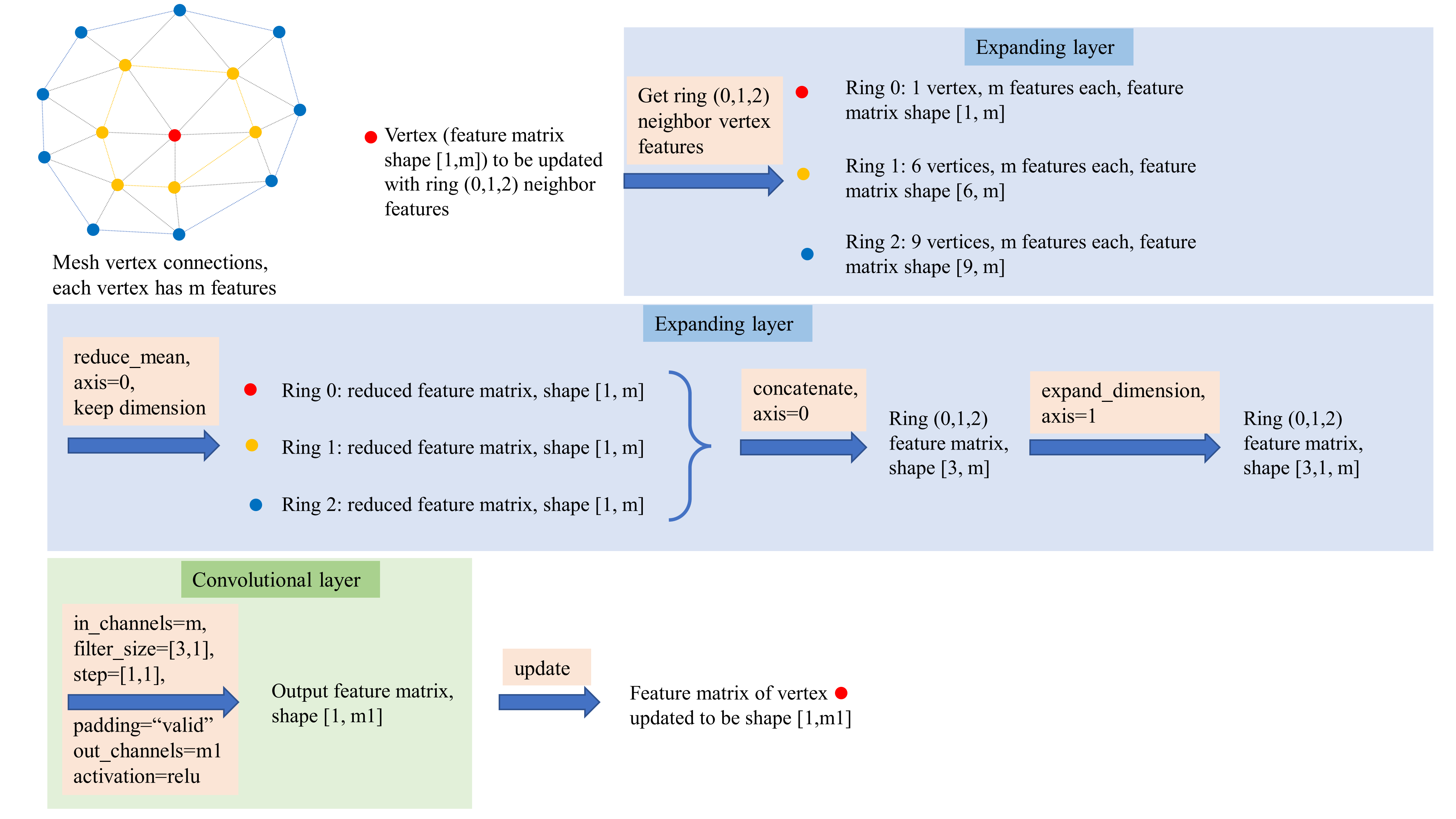}
	\caption{Illustration of operations in expanding layer and convolution layer}
	\label{fig:opers}
\end{figure*}

The operations of expanding layer and convolution layer were illustrated in Figure \ref{fig:opers}. All other vertices besides the red one will be set as center vertex and subjected to the same operation pipelines to update their feature matrices. The updated feature matrix contain learned information (by CNNs) from ring neighbor vertices, thus is able to extract local area features automatically. Once all vertices in a mesh updated their feature matrices, next iteration of expanding and convolution operations can be scheduled to extract deeper features.

Based on expanding layers and convolution layers, 6 kinds of networks were constructed, of which a network without expanding layer was set as baseline, as in Table \ref{tab:t1}.

\begin{table*}[ht]
	\centering
	\caption{Proposed architecture of mesh CNNs}
	\label{tab:t1}
	\setlength{\tabcolsep}{1mm}
	{
		\begin{tabular}{|c|c|c|c|c|c|}
			\hline
			Baseline & A  & B  & C  &    D       &     E      \\ 
			\hline
			\multicolumn{6}{|c|}{Input $\left[1, n, m\right]$ matrix, of which $batch\_size=1, vertical\_num=n, feature\_num=m$}                 \\ 
			\hline
			\makecell{$\left[ conv \left(3 \times 1 \right) - 16 \right]$ \\ $\times 2 $} 
			& 
			\makecell{$ \begin{bmatrix}	Expand \left(0,1,2 \right) \\ conv \left( 3 \times 1 \right) - 16 \end{bmatrix} $ \\ $\times 2 $} 
			& 
			\makecell{$ \begin{bmatrix}
				Expand \left(0,2,4 \right) \\ conv \left( 3 \times 1 \right) - 16 \end{bmatrix} $ \\ $\times 2 $}  
			& 
			\makecell{$ \begin{bmatrix}
				Expand \left(0,4,8 \right) \\ conv \left( 3 \times 1 \right) - 16 \end{bmatrix} $ \\ $\times 2 $}  
			&
			\multicolumn{2}{c|}{
				\makecell{$ \begin{bmatrix}
					Expand \left(0,1,2 \right) \\ conv \left( 3 \times 1 \right) - 16 \end{bmatrix} $ \\ $\times 2 $}
			} \\ 
			\hline
			\makecell{$\left[ conv \left(3 \times 1 \right) - 32 \right]$ \\ $\times 3 $} 
			& 
			\makecell{$ \begin{bmatrix}	Expand \left(0,1,2 \right) \\ conv \left( 3 \times 1 \right) - 32 \end{bmatrix} $ \\ $\times 3 $} 
			& 
			\makecell{$ \begin{bmatrix}
				Expand \left(0,2,4 \right) \\ conv \left( 3 \times 1 \right) - 32 \end{bmatrix} $ \\ $\times 3 $}  
			& 
			\makecell{$ \begin{bmatrix}
				Expand \left(0,4,8 \right) \\ conv \left( 3 \times 1 \right) - 32 \end{bmatrix} $ \\ $\times 3 $}  
			&
			\multicolumn{2}{c|}{
				\makecell{$ \begin{bmatrix}
					Expand \left(0,2,4 \right) \\ conv \left( 3 \times 1 \right) - 32 \end{bmatrix} $ \\ $\times 3 $}
			} \\ 
			\hline
			\makecell{$\left[ conv \left(3 \times 1 \right) - 64 \right]$ \\ $\times 3 $} 
			& 
			\makecell{$ \begin{bmatrix}	Expand \left(0,1,2 \right) \\ conv \left( 3 \times 1 \right) - 64 \end{bmatrix} $ \\ $\times 3 $} 
			& 
			\makecell{$ \begin{bmatrix}
				Expand \left(0,2,4 \right) \\ conv \left( 3 \times 1 \right) - 64 \end{bmatrix} $ \\ $\times 3 $}  
			& 
			\makecell{$ \begin{bmatrix}
				Expand \left(0,4,8 \right) \\ conv \left( 3 \times 1 \right) - 64 \end{bmatrix} $ \\ $\times 3 $}  
			&
			\multicolumn{2}{c|}{
				\makecell{$ \begin{bmatrix}
					Expand \left(0,4,8 \right) \\ conv \left( 3 \times 1 \right) - 64 \end{bmatrix} $ \\ $\times 3 $}
			} \\ 
			\hline
			\makecell{$\left[ conv \left(3 \times 1 \right) - 32 \right]$ \\ $\times 3 $} 
			& 
			\makecell{$ \begin{bmatrix}	Expand \left(0,1,2 \right) \\ conv \left( 3 \times 1 \right) - 32 \end{bmatrix} $ \\ $\times 3 $} 
			& 
			\makecell{$ \begin{bmatrix}
				Expand \left(0,2,4 \right) \\ conv \left( 3 \times 1 \right) - 32 \end{bmatrix} $ \\ $\times 3 $}  
			& 
			\makecell{$ \begin{bmatrix}
				Expand \left(0,4,8 \right) \\ conv \left( 3 \times 1 \right) - 32 \end{bmatrix} $ \\ $\times 3 $}  
			&
			\multicolumn{2}{c|}{
				\makecell{$ \begin{bmatrix}
					Expand \left(0,2,4 \right) \\ conv \left( 3 \times 1 \right) - 32 \end{bmatrix} $ \\ $\times 3 $}
			} \\ 
			\hline
			\makecell{$\left[ conv \left(3 \times 1 \right) - 16 \right]$ \\ $\times 2 $} 
			& 
			\makecell{$ \begin{bmatrix}	Expand \left(0,1,2 \right) \\ conv \left( 3 \times 1 \right) - 16 \end{bmatrix} $ \\ $\times 2 $} 
			& 
			\makecell{$ \begin{bmatrix}
				Expand \left(0,2,4 \right) \\ conv \left( 3 \times 1 \right) - 16 \end{bmatrix} $ \\ $\times 2 $}  
			& 
			\makecell{$ \begin{bmatrix}
				Expand \left(0,4,8 \right) \\ conv \left( 3 \times 1 \right) - 16 \end{bmatrix} $ \\ $\times 2 $}  
			&
			\multicolumn{2}{c|}{
				\makecell{$ \begin{bmatrix}
					Expand \left(0,1,2 \right) \\ conv \left( 3 \times 1 \right) - 16 \end{bmatrix} $ \\ $\times 2 $}
			} \\ 
			\hline
			\makecell{$\left[ conv \left(3 \times 1 \right) - 4 \right]$ \\ $\times 2 $} 
			& 
			\makecell{$ \begin{bmatrix}	Expand \left(0,1,2 \right) \\ conv \left( 3 \times 1 \right) - 4 \end{bmatrix} $ \\ $\times 2 $} 
			& 
			\makecell{$ \begin{bmatrix}
				Expand \left(0,2,4 \right) \\ conv \left( 3 \times 1 \right) - 4 \end{bmatrix} $ \\ $\times 2 $}  
			& 
			\makecell{$ \begin{bmatrix}
				Expand \left(0,4,8 \right) \\ conv \left( 3 \times 1 \right) - 4 \end{bmatrix} $ \\ $\times 2 $}  
			&
			\makecell{$ \begin{bmatrix}
				Expand \left(0,1,2 \right) \\ conv \left( 3 \times 1 \right) - 4 \end{bmatrix} $ \\ $\times 2 $}
			&
			\makecell{FC-128 \\ FC-512}
			\\ 
			\hline
			\makecell{$\left[ conv \left(3 \times 1 \right) - 2 \right]$ \\ $\times 1 $} 
			& 
			\makecell{$ \begin{bmatrix}	Expand \left(0,1,2 \right) \\ conv \left( 3 \times 1 \right) - 2 \end{bmatrix} $ \\ $\times 1 $} 
			& 
			\makecell{$ \begin{bmatrix}
				Expand \left(0,2,4 \right) \\ conv \left( 3 \times 1 \right) - 2 \end{bmatrix} $ \\ $\times 1 $}  
			& 
			\makecell{$ \begin{bmatrix}
				Expand \left(0,4,8 \right) \\ conv \left( 3 \times 1 \right) - 2 \end{bmatrix} $ \\ $\times 1 $}  
			&
			\makecell{$ \begin{bmatrix}
				Expand \left(0,1,2 \right) \\ conv \left( 3 \times 1 \right) - 2 \end{bmatrix} $ \\ $\times 1 $}
			&
			\makecell{FC-128 \\ FC-2}
			\\ 
			\hline
			\multicolumn{6}{|c|}{Softmax}                 \\ \hline
		\end{tabular}
	}
\end{table*}

\section{Experiment}

We collected 1445 dental meshes, the count of vertex in each mesh was controlled to be less than 6000, due to the limitation of our computation resources on a 8GB graphics card. Meshes were saved as .obj format, recording the vertical coordinates and its RGB color values, as well as vertex connections that composition triangle facets. Meshes were annotated that the teeth margin vertexes and inter-teeth vertexes were brushed to be red (Figure \ref{fig:annos}), that is, labeled as positive $\left(value = 1\right)$. All other vertexes were labeled as negative $\left(value = 0\right)$. 

\begin{table*}[ht]
	\centering
	\caption{Experiment results of trained mesh CNNs on validation dataset}
	\label{tab:t2}
	\setlength{\tabcolsep}{1mm}
	{
		\begin{tabular}{|c|c|c|c|c|c|}
			\hline
			\multirow{2}{*}{Classifier} & \multirow{2}{*}{Network} & \multirow{2}{*}{Training Features} & \multicolumn{3}{c|}{Validation}                  \\ \cline{4-6} 
			&                          &                                                                                & Accuracy       & Precision      & Recall         \\ \hline
			1                           & Baseline                 & Curvatures ($K_{max}, K_{min}, Mean, Guassian$), $d$                                         & 0.766          & 0.670          & 0.455          \\ \hline
			2                           & A                        & Curvatures ($K_{max}, K_{min}, Mean, Guassian$), $d$                                         & 0.863          & 0.786          & 0.756          \\ \hline
			3                           & B                        & Curvatures ($K_{max}, K_{min}, Mean, Guassian$), $d$                                         & 0.870          & 0.784          & $\mathbf{0.790}$ \\ \hline
			4                           & C                        & Curvatures ($K_{max}, K_{min}, Mean, Guassian$), $d$                                         & 0.857          & 0.784          & 0.733          \\ \hline
			5                           & E                        & Curvatures ($K_{max}, K_{min}, Mean, Guassian$), $d$                                        & 0.874          & 0.794          & 0.787          \\ \hline
			6                           & D                        & Curvatures ($K_{max}, K_{min}, Mean, Guassian$), $d$                                         & $\mathbf{0.877}$ & $\mathbf{0.799}$ & 0.789          \\ \hline
			7                           & D                        & $x, y, z$                                                                        & 0.807          & 0.695          & 0.649          \\ \hline
			8                           & D                        & $x, y, z$, Curvatures ($K_{max}, K_{min}, Mean, Guassian$), $d$                                & 0.866          & 0.783          & 0.773          \\ \hline
		\end{tabular}
	}
\end{table*}

For each vertex, 8 features were collected, including coordinates ($x, y, z$), curvatures ($K_{max}, K_{min}, Mean, Guassian$) and the mean neighbor distance ($d$). Curvatures ($K_{max}, K_{min}, Mean, Guassian$) are essential for description of the 3-dimensional morphology of meshes, and both curvatures and distance have shift-invariance and rotate-invariance properties. The mean neighbor distance ($d$) of a vertex $v$ was defined here as mean distance between $v$ and its ring 1 neighbor vertexes:
\begin{equation}
d=\frac{1}{n}\sum_{v_{i} \in \boldsymbol{N}_{1}\left ( v \right )}\sqrt{\left ( x-x_{i} \right )^{2}+\left ( y-y_{i} \right )^{2}+\left ( z-z_{i} \right )^{2}}
\end{equation}
where $x,y,z$ are coordinates of vertex $v$, and $x_{i},y_{i},z_{i}$ are coordinate of vertex $v_{i}$. The dataset was divided to be train dataset of 1156 meshes, validation dataset of 145 meshes and test dataset of 144 meshes.

\begin{figure*}[htbp]
	\centering
	\includegraphics[scale=0.36]{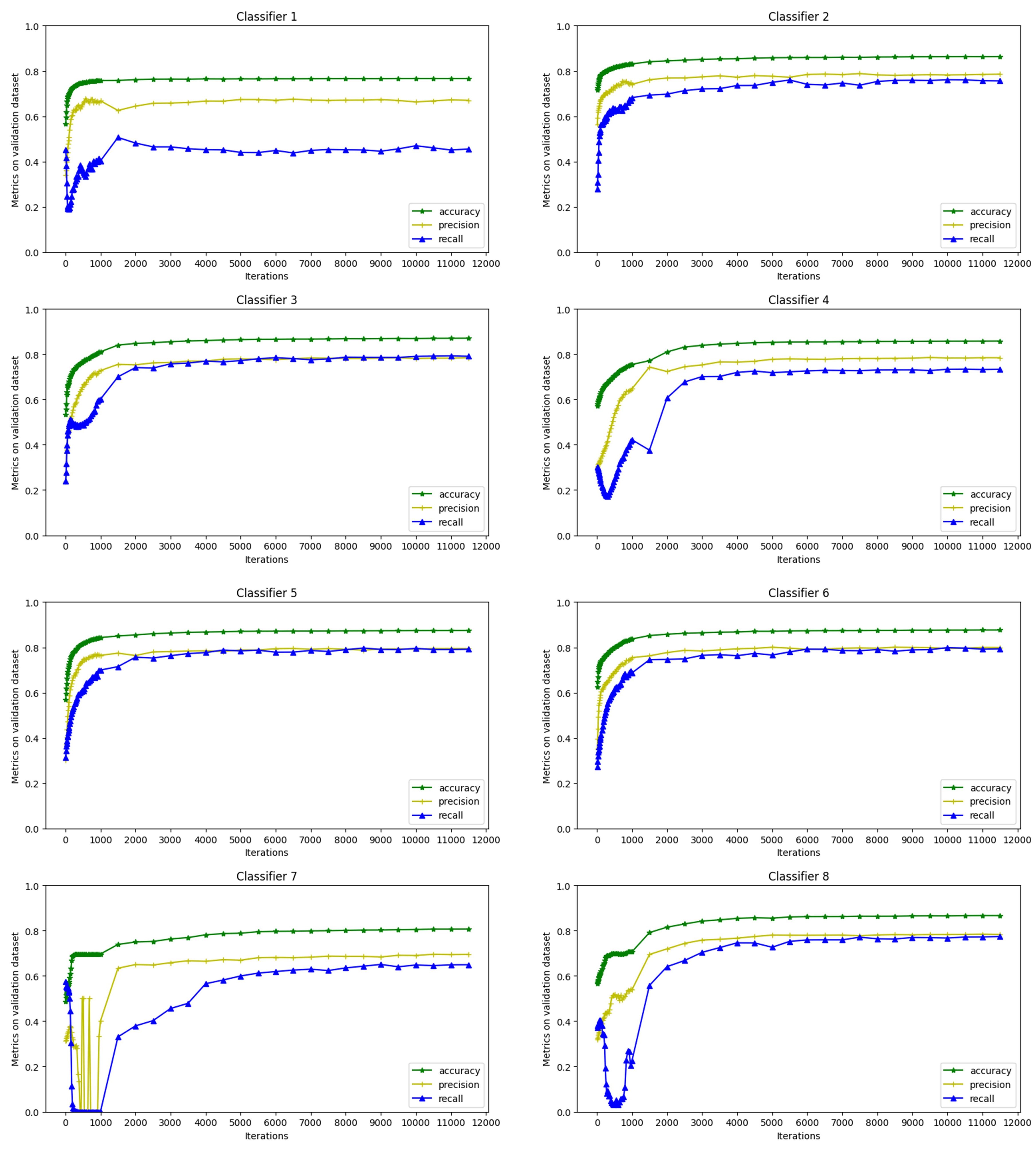}
	\caption{Accuracy, precision and recall on validation dataset during train iterations}
	\label{fig:val}
\end{figure*}

In this research, networks should be trained by labeled data before they have ability to do classifications. A trained network here was specified as a Classifier. Networks shown in Table \ref{tab:t1} were trained with different input features, as shown in Table \ref{tab:t2}, where classifier 1 to 6 were designed to test the performance of different network architectures, while 6 to 8 were designed to test the influence of different combination of input features on the outcomes. The loss function was defined upon weighted cross entropy with logits:

\begin{equation}
\begin{aligned}
Loss=&-\frac{1}{n}\sum{posWeight \ast target \ast log \left ( sigmoid \left ( logits \right ) \right )} \\
&-\frac{1}{n}\sum{\left ( 1-targets \right ) \ast \left ( 1-sigmoid \left ( logits \right ) \right )}
\end{aligned}
\end{equation}
where the $posWeight$ was set to be 3.0, and $Gradient Descent Optimizer$ was used to minimize the $Loss$, with initial $learning rate$ of 0.01, and decent to be 0.003 after 5000 train iterations, and 0.001 after 10000. This work was implemented on Tensorflow, and a total of 11500 train steps were performed.
During training, the network was evaluated on validation dataset after every 500 iterations. Accuracy, precision and recall were calculated:
\begin{equation}
Accuracy=\frac{TP+TN}{P+N}
\end{equation}
\begin{equation}
Precision=\frac{TP}{TP+FP}
\end{equation}
\begin{equation}
Recall=\frac{TP}{TP+FN}
\end{equation}
where $P$ represent positive vertexes, i.e. vertexes that were predicted to be teeth margin vertexes, $N$ present negative vertexes, i.e. background vertexes. $TP$ represent true positive vertexes, $TN$ true negative vertexes, $FP$ false negative vertexes, and $FN$ false negative vertexes. Validation results were illustrated in Figure \ref{fig:val}. Accuracy, precision and recall were all increased rapidly and reached a high level after about 4000 iterations, except that the recall of classifier 1 with baseline architecture dropped during the initial increase of iterations.

The final results were also shown in Table \ref{tab:t2}, of which the network with best validation results was chosen to be tested on test dataset.

\begin{figure*}[htbp]
	\centering
	\includegraphics[scale=0.19]{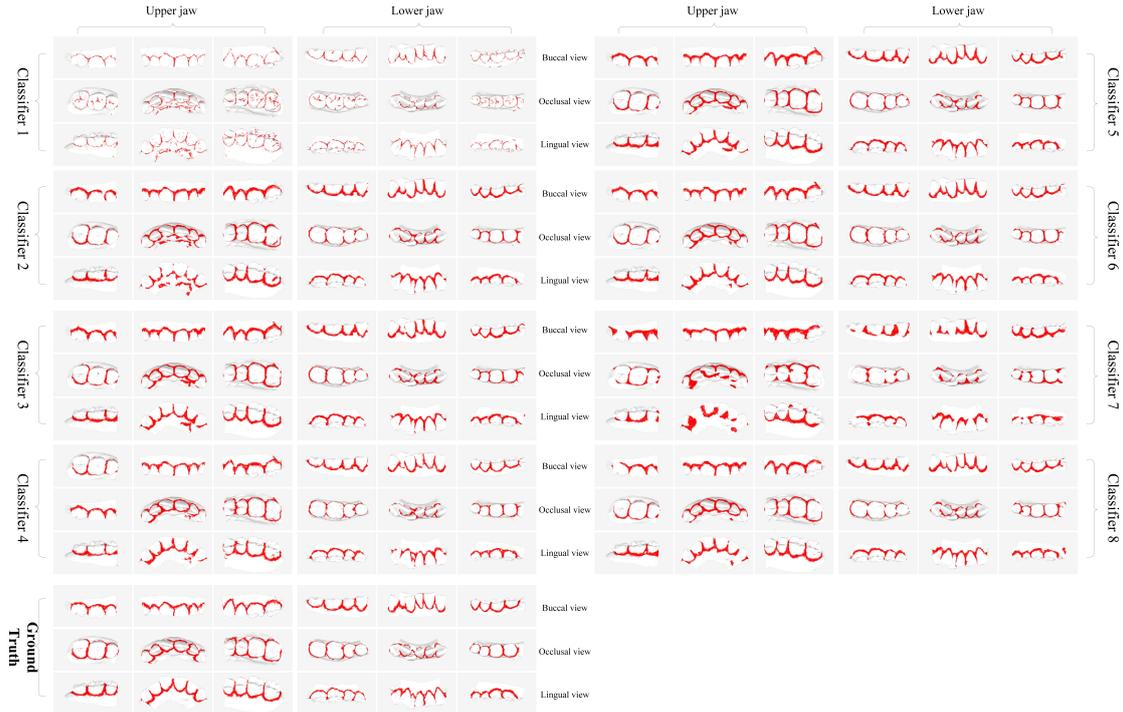}
	\caption{Ground truth and predicted margin vertexes (red) by classifier 1-8 on test dataset}
	\label{fig:fig1}
\end{figure*}

Predicted margin vertexes were shown in Figure \ref{fig:fig1}. Classifier 1, baseline network without expanding layer, worked like threshold segmentation, where areas with high curvature value were predicted as margin vertexes, producing false positive annotations in pit and fissure areas on occlusal surfaces and palatal wrinkle areas on upper jaws. Classifier 7 received raw vertex coordinates as input, roughly outlined the margin areas with obvious patch areas (false positive predictions) and gaps (false negative predictions). Gingiva margin areas were well predicted by classifier 2, 3, 4, 5, 6, 8, although there tend to be few false negative predictions in flat margin areas, and false positive predictions in rugged non-margin areas.

The accuracy, precision and recall of classifier 1-8 on validation dataset was shown in Table \ref{tab:t2}. Without expanding layers, the baseline reached the lowest accuracy of 0.766, as well as the lowest precision of 0.670 and recall of 0.455. The order of accuracy between networks were: $C < A < B < E < D$, and between features were: $coordinates < coordinates + Curvatures + d < Curvatures + d$. Classifier 6, which is network $D$ trained with curvatures $\left(K_{max}, K_{min}, Mean, Guassian\right)$ and mean neighbor distances ($d$), achieved highest accuracy and precision among all the 8 classifiers. The recall of classifier 6 also reach 0.789, which is only 0.001 lower than the highest one achieved by classifier 3. The classifier 7, taking raw coordinates as train features, got lowest accuracy among classifier 2-8, but still higher than baseline.

Since classifier 6 performed best among classifier 1-8, it was selected to predict vertex labels on test dataset. Results shown that the accuracy, precision and recall were 0.877, 0.782 and 0.804, respectively, which was close to the values predicted on the validation dataset (by classifier 6).

\section{Discussion}
Expanding layer here align vertex features with its neighbor vertex features, and then subjected to convolution layer, multiplied by convolution kernels, to expand the receptive field. The linkage between vertexes defined by the mesh edges and facets were rarely considered in previous studies, which is very important to revel the topological structure of mesh. However, the number of adjacent vertexes varied from point to point, which could not be presented by a matrix as in 2 dimensional images, thus convolution was not available. In this work, adjacent relationship was presented as ring neighbor vertexes linkages, and each ring neighbor vertexes’ features were reduced to mean, equivalent to features of one vertex in number. After expanded the input features to specified number of ring neighbors, a new feature matrix with adjacent feature information was constructed. Then the expanded feature matrix passed through a conv layer to learn the reduced ring neighbor feature pattern, and output feature matrix with the same shape as initial input, which could pass though expanding layer plus conv layer iteratively to enlarge its receptive field.

Expand to further ring neighbors will expand receptive field efficiently, but may miss some nearby information. Classifier 2 with expanding to ring 0, 2 and 4 performed better than classifier 1 (expanding to ring 0, 1 and 2) and classifier 3 (expanding to ring 0, 4 and 8), implies moderate expanding is preferred. What’s more, a combination of different expanding distances in classifier 6 and 7 gained further improvement. After the last 3 expanding plus conv layer blocks replaced by fully connected layers, classifier 7 performed slightly inferior than classifier 6. This might be explained that fully connected layers didn’t concern local mesh vertex topological structures as expanding plus conv layers did. With regarding to the input mesh vertex features, it seems that the curvature and distance features are better than coordinates feature to train our mesh CNN. This may because curvatures and distances have shift-invariance and rotate-invariance properties, which will not be affected by the inconsistent of pose or position of the teeth scanned.

In this work, negative samples were much more than positive samples (teeth margin vertexes), although we brushed a thick rough line at teeth margin area to mark more vertexes as positive samples, the ratio was still about 1:3 for positive: negative. The loss function was defined upon weighted cross entropy with logits to balance the samples. If not weighted, this unbalanced distribution may affect the performances of our networks, with potential to assign all vertices with negative labels to achieve a local optimum. 

There are ways to shrinkage the thick rough teeth margin strip areas to be lines, such as grassfire algorithm, where a continuous gingival strip area without gaps is required. With classifier 6 in this study, there is still few gaps in flat areas. However, this kind of gaps can be easily filled interactivity with little manual labor.

Although convolution of mesh vertex features was implemented in this work, subsampling and upsampling were not covered, which is very important for semantic segmentation of mesh. The subsampling and upsampling will come down to remeshing, which will be investigated in our further work.

\section{Conclusion}

In this research, we constructed an end-to-end neural network which is capable to directly take mesh feature data and vertex connection information as input, and achieved good performance on teeth mesh vertex classification. The expanding layers designed to extract local vertex features can significantly improve the classification result, and the strategy of assembly of expanding layers with different reach sizes can be properly investigated to achieve better performance. Mesh vertex features with shift-invariance and rotate-invariance properties are preferred to train our mesh CNN.

\section*{acknowledgements}
This study was funded by the National Natural Science Foundation of China (No. 51705006), and open fund of Shanxi Province Key Laboratory of Oral Diseases Prevention and New Materials (KF2020-04).

\section*{Conflict of interest}
The authors declare that they have no conflict of interest.

%\bibliographystyle{wileynum}

%\bibliography{\jobname}
\end{document}